\documentclass{article}

\usepackage{spconf,amsmath,epsfig}

\usepackage{epsfig}
\usepackage{amssymb}
\usepackage{color}
\usepackage[table]{xcolor}
\usepackage{url}
\usepackage{tabularx}
\usepackage{multirow}
\usepackage{makecell}
\usepackage{array}
\usepackage{caption}

\let\OLDthebibliography\thebibliography
\renewcommand\thebibliography[1]{
  \OLDthebibliography{#1}
  \setlength{\parskip}{0pt}
  \setlength{\itemsep}{0pt plus 0.3ex}
}

\begin{document}\sloppy
\captionsetup{belowskip=-15pt}

\def\x{{\mathbf x}}
\def\L{{\cal L}}

\title{Leveraging Compressed Frame Sizes for Ultra-Fast Video Classification}
%
\name{Yuxing Han, Yunan Ding, Chen Ye Gan, Jiangtao Wen}
\address{}

\maketitle

\begin{abstract}
Classifying videos into distinct categories, such as Sport and Music Video, is crucial for multimedia understanding and retrieval, especially when an immense volume of video content is  being constantly generated. Traditional methods require video decompression to extract pixel-level features like color, texture, and motion, thereby increasing computational and storage demands. Moreover, these methods often suffer from performance degradation in low-quality videos. We present a novel approach that examines only the post-compression bitstream of a video to perform classification, eliminating the need for bitstream decoding. To validate our approach, we built a comprehensive data set comprising over 29,000 YouTube video clips, totaling 6,000 hours and spanning 11 distinct categories. Our evaluations indicate precision, accuracy, and recall rates consistently above 80\%, many exceeding 90\%, and some reaching 99\%. The algorithm operates approximately 15,000 times faster than real-time for 30fps videos, outperforming traditional Dynamic Time Warping (DTW) algorithm by seven orders of magnitude. 
\end{abstract}
\section{Introduction}
\label{sec:intro}

Video classification is fundamental for multimedia services, enabling functionalities such as content retrieval, recommendation, and optimized coding. Traditional techniques focus on analyzing video features such as color, texture, and motion, while recent approaches incorporate deep learning-based algorithms\cite{donahue2015long}\cite{tran2015learning}\cite{zolfaghari2018eco}.

A major drawback of current approaches is their dependency on pixel-domain features, resulting in computational and storage inefficiencies. This issue is exacerbated by the high-volume video uploads to platforms like YouTube and TikTok, where videos are typically compressed. Extracting pixel data from such compressed videos necessitates full decoding, leading to a storage increase ratio of up to 75:1 for a 1080p30 video compressed at 10 Mbps. Even with optimizations like down-sampling spatial and temporal resolutions and specialized low-memory-footprint classification techniques \cite{bhardwaj2019efficient}\cite{kondratyuk2021movinets}, the classification of the 30,000 hours of video uploaded to YouTube every hour would necessitate a method operating thousands of times faster than real-time, while consuming hundreds of times more storage. Furthermore, these techniques often struggle in classifying low-quality videos and raise privacy issues, as decryption is needed. Some videos, due to DRM policies, cannot even be decrypted during transmission.

We introduce a novel direction for video classification that does not rely on pixel domain information. Instead, we use the sequence of video frame sizes extracted from compressed bitstreams as input for a ResNet-based deep neural network, without the need for bitstream decoding or parsing. This approach leverages information captured by modern video compression algorithms, particularly advanced spatial and temporal prediction methods found in modern video coding standards such as H.264/AVC, H.265.HEVC and H.266/VVC. Optimized encoders also incorporate advanced rate control algorithms, which allocates and enforces bitrates across and within frames. At a high level, the complexity of individual frames determines the size of the reference frames; uniformity of the frames determines the size of the predicted frames; the frequency of scene changes is determined by factors such as camera movement rhythms. 

Using solely this compressed bitstream, our approach significantly reduces computational and storage demands, maintains precision, accuracy, and recall rates of at least 80\%, often exceeding 90\%, while ensuring privacy and security.

\section{Related Work}

Video classification using the time sequence of the sizes of video frames after 
compression (in encoding order) is a typical
Time Series Classification (TSC) problem \cite{bagnall2017great}.
We define the sizes of compressed frames in bits in encoding order 
as a time series $X = {[x_1,x_2,x_3,...,x_T]} \in \mathbb{R}^T$, where T represents the length of the time series. Consequently, video classification can be formulated as a mapping between $X$ and a one-hot label vector $Y$.

Traditional TSC algorithms often use distance-based metrics with 
a $k$-nearest neighbor classifier ($K$-NN). Such algorithms 
include Dynamic Time Warping (DTW) \cite{berndt1994using}, Weighted DTW  
\cite{jeong2011weighted}, Move–split–merge \cite{stefan2012move}, Complexity invariant 
distance \cite{batista2014cid}, Derivative DTW \cite{gorecki2013using}, Derivative 
transform distance \cite{gorecki2014no}, Elastic ensemble \cite{lines2015time}, etc., 
where DTW is often used as a baseline \cite{bagnall2017great}. 
All such methods take the entire time 
series as input and are computationally intensive. Even following the reduced
complexity TSC algorithm from Rodríguez et al. \cite{rodriguez2004support}, 
TSC algorithms are in general, still highly time-consuming. 

In recent years, deep learning based TSC has been studied broadly. A 
study by Fawaz et al.\cite{ismail2019deep} evaluated the performances of nine deep learning models for TSC on the univariate UCR/UEA archive benchmark \cite{c.dataset1} and 12 Multivariate Time Series (MTS) data sets \cite{c.dataset2}. Among them the most recent and best performing are Encoder \cite{serra2018towards} and Residual Network (ResNet) \cite{wang2017time}, respectively. 

Based on the above studies, we used DTW as a baseline for traditional TSC algorithms and ResNet as the representative structure for deep learning based algorithms.

\section{Proposed Method}

\begin{figure}[t]
    \centering
    \includegraphics[width=0.98\columnwidth]{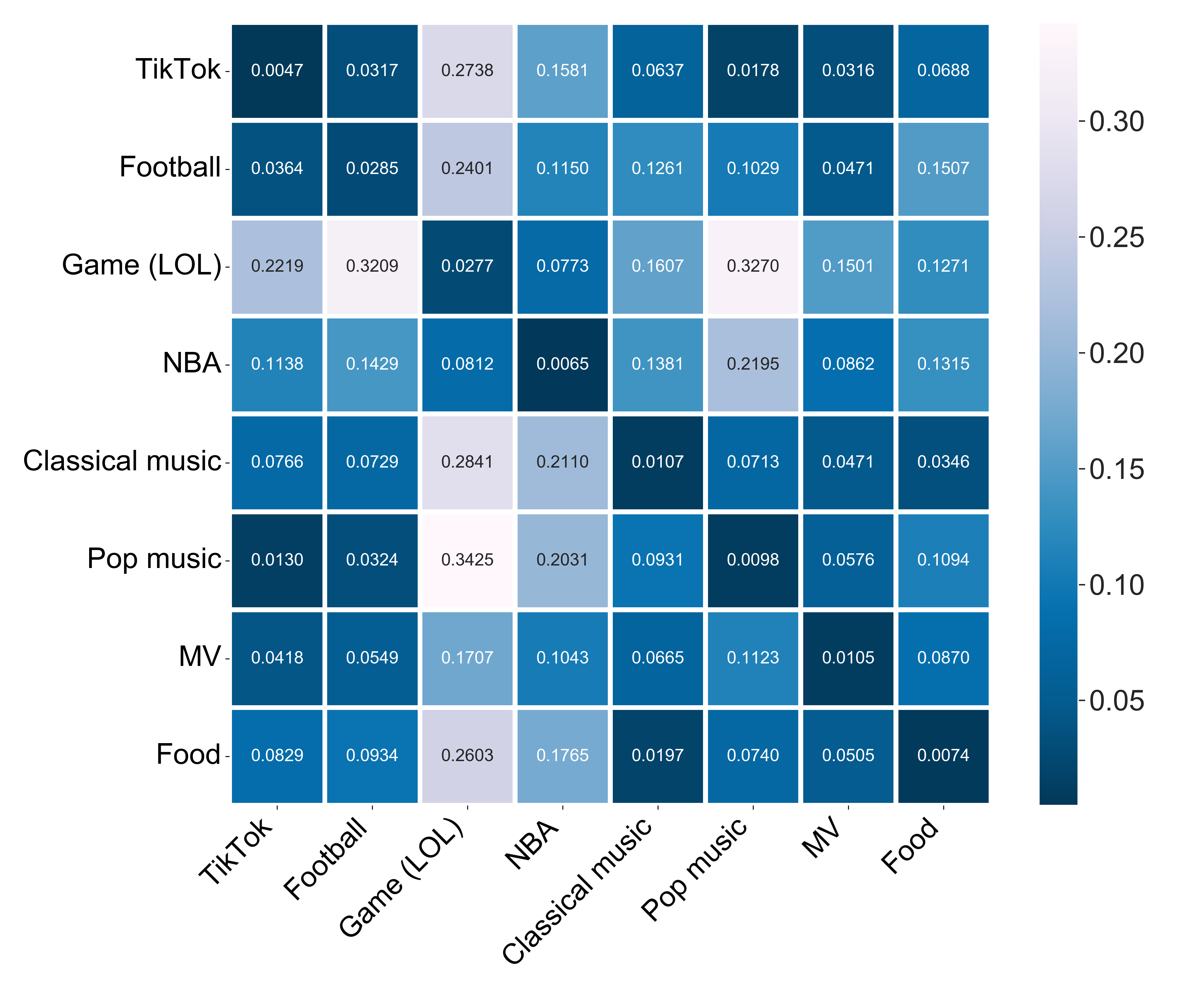}
    \caption{Kullback-Leibler Divergence}
    \label{fig:Kullback-Leibler Diverge}
\end{figure}

We propose that videos across various categories exhibit unique style and editing traits which manifests in discernible bitstream distributions, entropy-based motion vectors, and bitrate allocations when processed by a rate-distortion optimized encoder, which could aid in the classification of stylistically varied content, such as NBA vs. Football or Pop vs. Classical concerts, or content from different social media influencers. This effect is even more pronounced for social network videos that use preset filters and special effects. 

\subsection{Correlation between Transcoded Bitstreams}

In our experiment, we standardized all clips to 1.5Mbps bitrate using FFmpeg’s H.264/AVC encoder. We then calculated the inter- and intra-class Kullback-Leibler Divergence (KLD) for 8 sub-classes within our data set, namely NBA, TikTok, Football, Classical Concerts, Pop Concerts, Gaming (League of Legends), Music Videos, and Culinary Exploration. As Figure \ref{fig:Kullback-Leibler Diverge} illustrates, intra-class KLD values are typically orders of magnitude lower than inter-class KLD values. This result indicates high correlation between the frame size time series of bitstreams with various with encoding settings, suggesting that 1) post-compression frame sizes, reflecting bitrate variations, is useful in video classification; and 2) a single model might be capable of classifying videos across a spectrum of bitrates.

\subsection{Implementation of Time Series Classifier}

\begin{figure}[t]
    \centering
    \includegraphics[width=0.98\columnwidth]{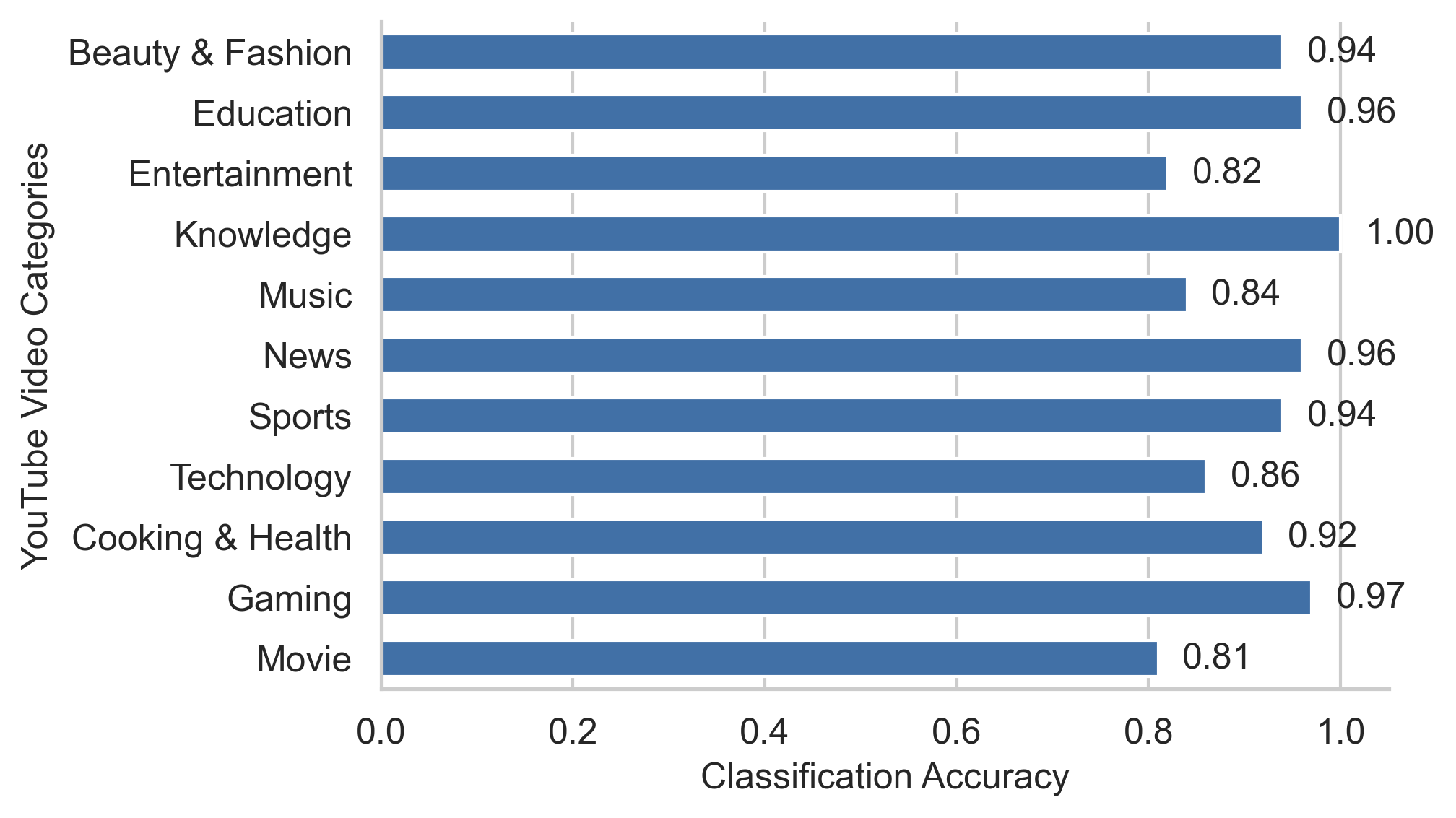}
    \caption{Effectiveness on broad video categories on YouTube}
    \label{fig:11_channel}
\end{figure}

\begin{figure}[t]
    \centering
    \includegraphics[width=0.98\columnwidth]{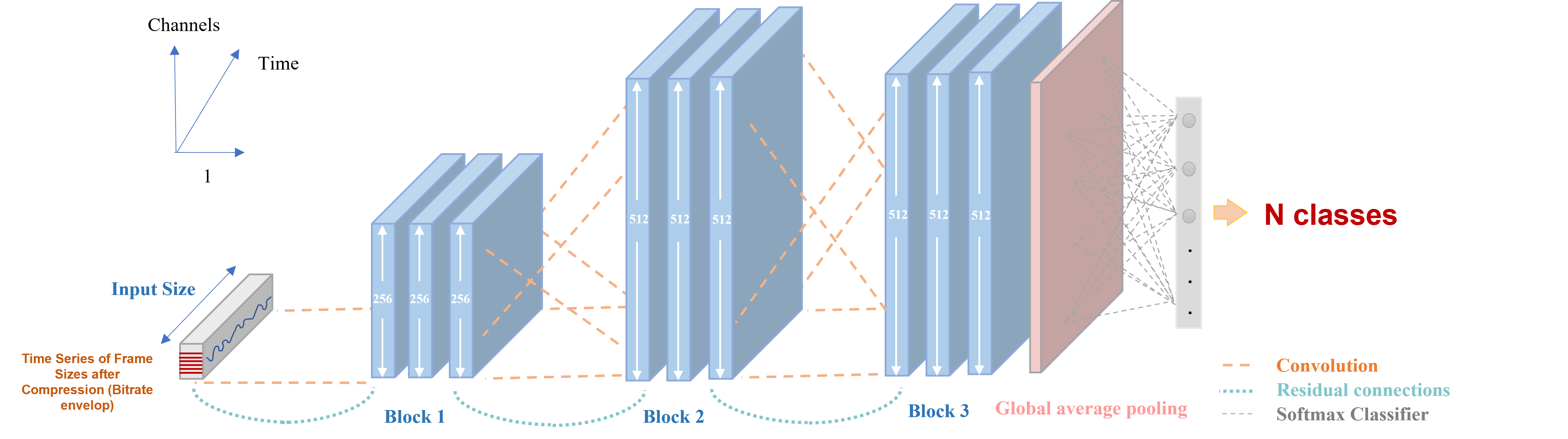}
    \caption{Residual architecture classifier based on \cite{wang2017time}}
    \label{fig:ResNet}
\end{figure}

To explore the potential of bitrate time series for video categorization, we opted for the established ResNet classifier \cite{he2016deep}, depicted in Figure \ref{fig:ResNet}.

\textbf{Model architecture.} The ResNet based classifier mainly consisted of $3$ residual blocks, which are used to extract features from the input time series. 
To describe the process of a classifier predicting the class of a given input X in this supervised learning task, the equation can be expressed as:
\begin{eqnarray}
\hat{y} &=& f_L(\theta_L, x) \nonumber \\
        &=& f_{L-1}(\theta_{L-1}, f_{L-2}(\theta_{L-2}, \dots, \nonumber \\
        && \qquad f_1(\theta_1, x + \mathcal{F}(x)))
\end{eqnarray}

\noindent where $\hat{y}$ is the predicted output, $f_{L}$ is the composite function (convolution, batch normalization, activation) applied at the $L$-th layer, $\theta_{L}$ denotes the set of parameters at the $L$-th layer, and $\mathcal{F}(x)$ denotes the shortcut connection in each residual block. The final output $\hat{y}$ of the ResNet model is produced by a global average pooling layer and a softmax classifier. The global average pooling layer generates feature maps of $\hat{y}$  for different video categories in this classification task. Then the softmax classifier maps the feature vector to a probability distribution over the output classes. Residual connection between convolutional layers effectively avoids vanishing gradient in the training process \cite{he2016deep}. The number of filters for each convolutional blocks are set to 256, 512, 512 respectively.

\textbf{Model Training Specifics.} We used fixed parameter initialization and an auto-adjusted learning rate, with the factor at 0.5 and the patience level at 40. To avoid overfitting, an early stop strategy was adopted with the patience set to 80. The adam optimizer was used for all contrast experiments.

For this multi-classification task, categorical cross-entropy loss function is used as the supervised training loss, which is written as

\begin{equation} \label{eq:cross-loss} \mathrm{Loss}(\mathbf{y}, \mathbf{\hat{y}}) = -\sum_{i=1}^{C} y_i \log(\hat{y}_i) \end{equation} 

\noindent where $y_i$ denotes one-hot encoding for the corresponding video category. $C$ denotes the total number of categories. Equation \ref{eq:cross-loss} aims to maximize the likelihood of predicting true labels given model parameters.

\section{Experimental Results}

\subsection{Data Preparation}
Existing data sets, such as YouTube-8M \cite{abu2016youtube}, 
Activity Net \cite{caba2015activitynet}, and Net Sports-1M \cite{karpathy2014large}, primarily focus on short-form content, often limiting clips to less than 5 seconds. 

We curated a comprehensive data set, sourced through keyword search on YouTube, comprising 29,142 video segments, each containing at least 3,000 frames, across 11 broad YouTube categories: Movies, Entertainment, Cooking \& Health, Gaming, Technology, Knowledge, Music, Sports, Beauty \& Fashion, News, and Education, and their corresponding bitstreams compressed using different encoding settings or downloaded from YouTube (“entropy coded covers”), collectively exceeding 6,000 hours in duration. 

Each category holds a minimum of 400 clips, each clip containing at least 3,000 frames. The clips vary in spatial resolution, from 298x480 for TikTok videos to 720p and 1080p for others, and are consistent in frame rate (either 30fps or 60fps) within their respective category.

Clips in some categories overlap conceptually—TikTok might also contain NBA or Dance clips—to test the algorithm's ability to discern “main characteristics” of the content. A viral NBA-themed clip on Tiktok usually follows a certain recognizably TikTok style, as a result of TikTok filters, editing tools, and various social marketing “rules” which makes it distinctively different from standard NBA broadcasts. We aim to distinguish a live broadcast NBA clip from the same clip in TikTok style.

\setlength{\tabcolsep}{4pt}

\begin{table*}[t]
    \centering
    \resizebox{0.88\textwidth}{!}{
    \color[HTML]{333333}\begin{tabular}{c|cccc|cccc|cccc|c}
        \hline
        \hline
        \multirow{3}{*}{Metric (\%) } & \multicolumn{8}{c|}{Constant Bitrate (kbps)}                     & \multicolumn{4}{c|}{Average Bitrate (kbps)}       &  Training \\ 
                                      & \multicolumn{4}{c|}{without B-frames} & \multicolumn{4}{c|}{with B-frames} & \multicolumn{4}{c|}{with B-frames}      &  Data Set \\  
                                      & 800 & 1000 & 1200 & 1500 & 800 & 1000 & 1200 & 1500              & 800 & 1000 & 1200 & 1500                          &  (kbps)   \\  
        \hline
        \hline

        Performance                   & {\color[HTML]{FF0000} 84.49} & 80.99 & 70.50 & 56.61 & {\color[HTML]{FF0000} 85.11} & 79.66 & 67.16 & 58.17 & {\color[HTML]{FE0000} 86.87} & 85.35 & 79.39 & 70.15 & \multirow{3}{*}{800} \\  
        Accuracy                      & {\color[HTML]{FF0000} 83.79} & 74.01 & 46.16 & 20.67 & {\color[HTML]{FF0000} 85.27} & 74.63 & 42.20 & 21.16 & {\color[HTML]{FE0000} 87.62} & 83.54 & 72.40 & 52.35 &                      \\  
        Recall                        & {\color[HTML]{FF0000} 85.33} & 76.33 & 50.64 & 28.39 & {\color[HTML]{FF0000} 85.72} & 77.27 & 48.34 & 29.31 & {\color[HTML]{FE0000} 88.01} & 84.79 & 75.90 & 56.91 &                      \\  

        \hline

        Performance                   & 78.63 & {\color[HTML]{FF0000} 83.47} & 80.89 & 68.80 & 85.16 & {\color[HTML]{FF0000} 87.93} & 84.80 & 66.10 & 86.82 & {\color[HTML]{FE0000} 87.98} & 85.55 & 78.12 & \multirow{3}{*}{1000} \\  
        Accuracy                      & 78.22 & {\color[HTML]{FF0000} 83.29} & 76.36 & 42.45 & 86.14 & {\color[HTML]{FF0000} 87.50} & 80.94 & 34.78 & {\color[HTML]{FE0000} 88.24} & 88.00 & 84.90 & 72.90 &                      \\  
        Recall                        & 78.47 & {\color[HTML]{FF0000} 84.44} & 78.62 & 48.08 & 86.67 & {\color[HTML]{FF0000} 88.14} & 82.67 & 42.65 & {\color[HTML]{FE0000} 88.65} & 88.48 & 86.00 & 76.65 &                      \\  

        \hline

        Performance                   & 69.78 & 77.83 & {\color[HTML]{FF0000} 82.18} & 79.59 & 79.19 & 84.65 & {\color[HTML]{FF0000} 85.76} & 79.34 & 69.78 & {\color[HTML]{FE0000} 75.19} & 74.86 & 72.25 & \multirow{3}{*}{1200} \\  
        Accuracy                      & 64.11 & 75.99 & {\color[HTML]{FF0000} 81.19} & 74.38 & 80.94 & 85.40 & {\color[HTML]{FF0000} 86.26} & 73.76 & 56.06 & 70.92 & {\color[HTML]{FE0000} 71.53} & 63.00 &                      \\  
        Recall                        & 64.25 & 76.32 & {\color[HTML]{FF0000} 82.81} & 77.94 & 81.74 & 86.48 & {\color[HTML]{FF0000} 87.47} & 76.69 & 59.07 & 74.62 & {\color[HTML]{FE0000} 75.56} & 68.84 &                      \\  

        \hline

        Performance                   & 59.41 & 72.15 & 78.54 & {\color[HTML]{FF0000} 84.11} & 66.07 & 73.66 & 81.19 & {\color[HTML]{FF0000} 84.55} & 72.67 & 78.55 & 84.76 & {\color[HTML]{FE0000} 85.23} & \multirow{3}{*}{1500} \\  
        Accuracy                      & 58.04 & 72.77 & 79.58 & {\color[HTML]{FF0000} 82.92} & 61.26 & 72.15 & 80.82 & {\color[HTML]{FF0000} 83.17} & 73.14 & 80.45 & {\color[HTML]{FE0000} 85.77} & 85.64 &                      \\  
        Recall                        & 55.06 & 72.01 & 80.28 & {\color[HTML]{FF0000} 84.41} & 60.65 & 71.06 & 80.70 & {\color[HTML]{FF0000} 83.24} & 73.68 & 80.56 & 86.30 & {\color[HTML]{FE0000} 86.35} &                      \\  
     
        \hline
        \hline

    \end{tabular}
    }
    \caption{Classification performance with ABR encoding at various train/test bitrates. }
    \label{tab:my-table(ABR)}
\end{table*}

\begin{figure}[t]
    \centering
    \includegraphics[width=0.95\columnwidth]{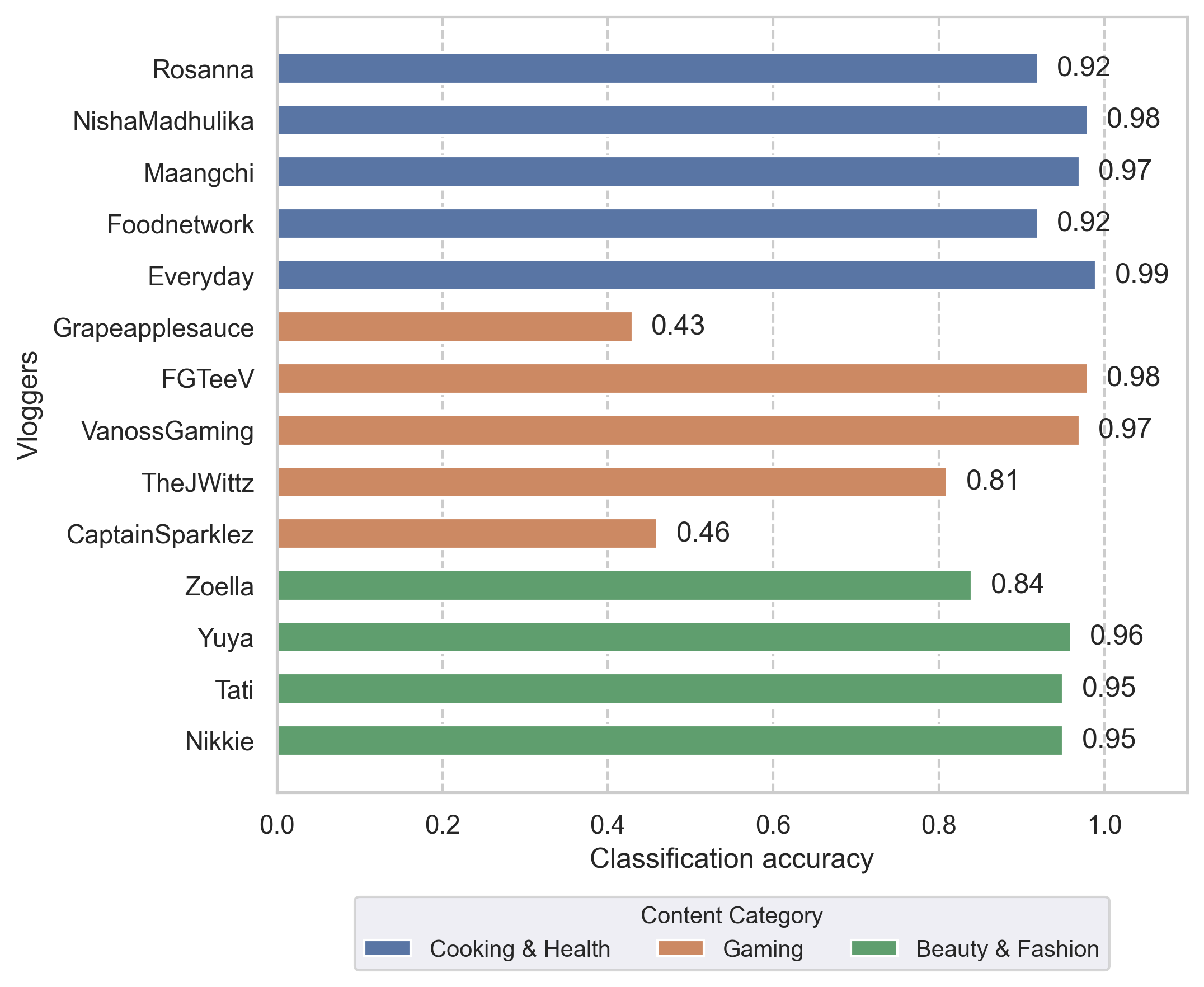}
    \caption{Evaluating ability to discriminate individual vloggers within a channel}
    \label{fig:same_channel}
\end{figure}

\subsection{Performance Evaluation and Applications}

We observed precision, accuracy, and recall rates consistently exceeding 80\%, as shown in Figure \ref{fig:11_channel}. The algorithm demonstrated particular sensitivity to distinct “editing styles”, including factors such as shot selection, camera angles, and movement. Remarkably, even when each video frame was represented by a singular numerical value, the classifier could effectively distinguish between diverse video categories such as NBA games, football matches, and classical or pop concerts. The proposed method was less effective in categories where editing styles are more diverse. However, it could  identify specific vloggers on YouTube as shown in Figure 
 \ref{fig:same_channel}.

\subsection{Classification Speed}
Our ResNet-based video classifier, using a single Nvidia A100 GPU, processed 1,818 test videos—each containing 3,000 frames—in less than 13 seconds. This corresponds to a real-time factor of approximately 15,000 for 30fps, within striking distance of the 30,000-to-1 ratio of total video duration
uploaded to YouTube per unit time.

In contrast, we evaluate the conventional non-deep-learning-based Dynamic
Time Warping (DTW) algorithm \cite{berndt1994using}. Using the same data set but restricting videos to short 30-frame clips, the DTW algorithm required an
extensive 27 hours for classification. Given DTW's computational complexity of $O(N^2)$, where $N$ is the time series length, the algorithm is approximately $7.5 \times 10^7$ times slower than the ResNet classifier, with an accuracy of less than 0.35. Although longer time series might improve DTW's performance, the algorithm's inefficiency makes it an impractical alternative.

\subsection{Impact of Input Length, Mismatched Training/Testing Bitrates, and Encoding Settings}

\begin{figure}[t]
    \centering
    \includegraphics[width=0.90\columnwidth]{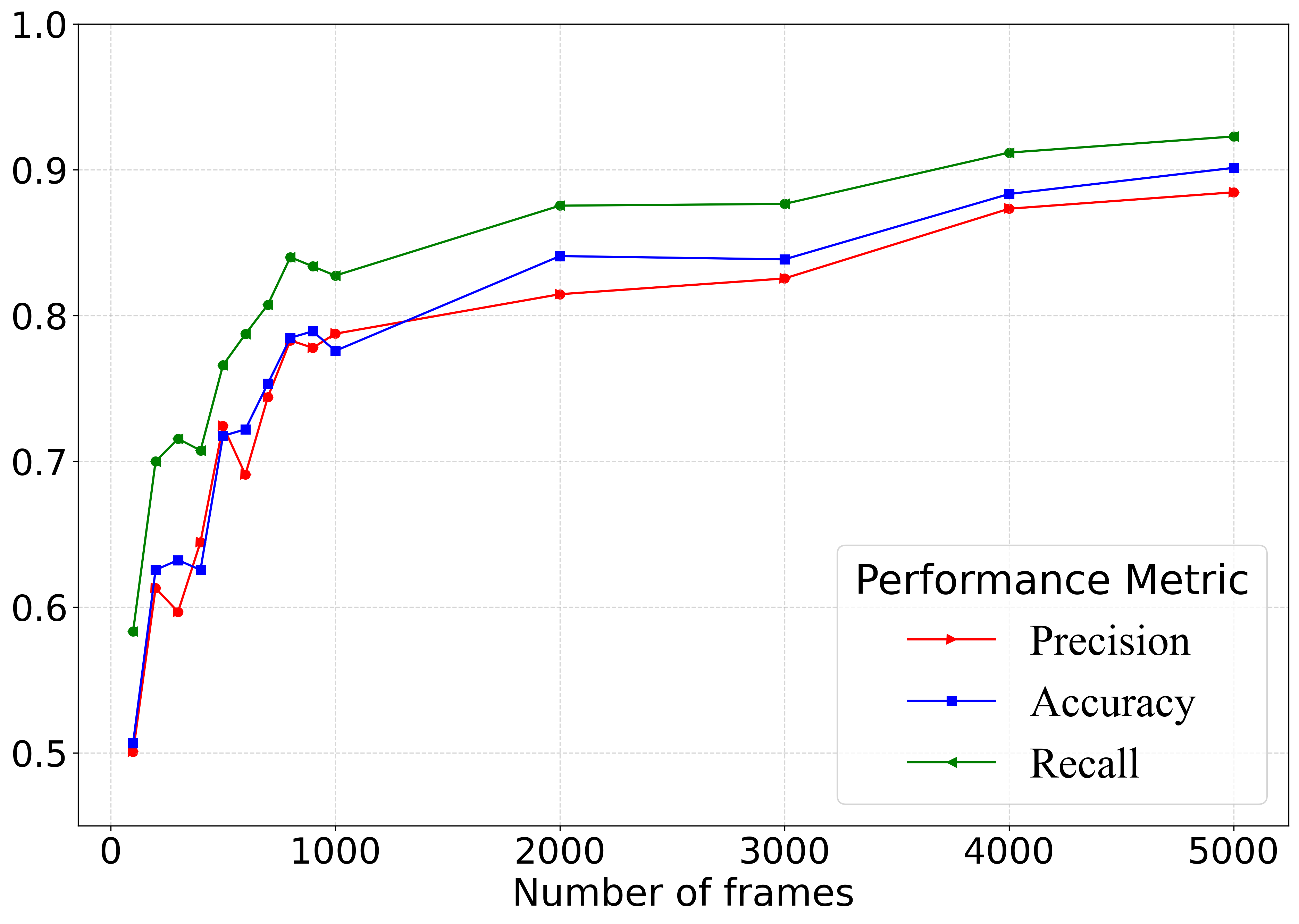}
    \caption{Classification performance as a function of input size.}
    \label{fig:FramePerm}
\end{figure}

We assess the classifier's performance as influenced by the number $N$ of frames utilized for model training and classification with a range of $N$ values: 120, 240, 360, 480, 600, 720, 840, 960, 1200, 2400, 3600, and 4800 frames. The corresponding classification outcomes for the Average Bitrate (ABR) scenario are depicted in Figure \ref{fig:FramePerm}. 

The results indicate that smaller $N$ values may compromise classification performance, likely due to localized content variations. As $N$ increases, classification efficacy improves until converging asymptotically to a stable state. 

Additionally, our approach remains robust when classifying bitstreams with bitrates that differ from the model's training bitrate. We trained models on 3,000-frame videos encoded at four different bitrates: 800kbps, 1000kbps, 1.2Mbps, and 1.5Mbps. Subsequently, we evaluated each model's performance on 3,000-frame videos encoded at all four bitrates. 

We also verified the robustness of our classifier across encoding settings, namely Average Bitrate (ABR) mode, Constant Bitrate (CBR) mode, and with and without B-frames. Table \ref{tab:my-table(ABR)} shows the classification performance for various combinations of training bitrates, testing bitrates, and encoding settings (ABR/CBR, B-frames). 

For videos encoded down to about 50\% of model training bitrate, e.g. classifying 800kbps videos using a model trained at 1.5Mbps, the classification remains effective. However, classification performance suffers when the input video bitrate is reduced to 1/3 of the model's training bitrate. 

\section{Conclusions and Discussions}

Traditionally, we’re cautioned against judging a book by its cover. However, much can be inferred about a book’s caliber from its cover—the paper quality, meticulous typesetting, and chosen color palette. Similarly, a video’s compression encoded bitstream is a treasure trove of information, enabling us to appraise a video by its ”bitstream cover” with impressive accuracy. This encoded bitstream, originates from a sophisticated encoding process and comprises of hundreds, if not thousands, of data points, makes it a prime target for deep learning to extract insights. 

In our experiments, we employed a straightforward classifier that utilizes the time series of compressed video frame sizes as input, which is readily accessible through various means such as byte-aligned headers or Network Abstraction Layer (NAL) packets without decoding. This design simplifies computational complexity and enhances data privacy. Remarkably, NAL packet-based analysis could even function with encrypted content, positioning our technique as a scalable solution for network carriers. 

We demonstrate robust classification capabilities (minimum of 80\% accuracy) in classifying 11 broad categories and 39 specific vloggers in 29,142 video segments, including those that overlap in content, and across various bitstream encoding settings (ABR, CBR, use of B-frames), obtaining similar performances. Furthermore, it processes 30fps videos approximately 15,000 times faster than real-time, outperforming the traditional DTW algorithm by seven orders of magnitude, revealing DTW's prohibitively high time complexity. 

Current limitations include the high number of frames (N$>$2000) required for training and classification to achieve accuracies above 80\%. For 30fps videos, this equates to a delay of about $2000/30 \approx 67s$ for real-time classifications. Additionally, the approach struggles with videos of similar editing styles, as seen in Gaming videos, where the predominant style is recorded gameplay accompanied by a floating head commentary, as well as differentiating certain finer nuances, such as discerning between NBA clips featuring different basketball stars. The maximum number of classes this method can distinguish remains to be established. 

This papers marks the first attempt at classifying video using compressed bitstreams. We have open sourced our model to encourage subsequent research on better performing models that do not require pixel-level decoding.


\bibliographystyle{IEEEbib}
\bibliography{refs}

\end{document}